\pdfoutput=1

\documentclass[11pt]{article}

\usepackage{naacl2021}

\usepackage{times}
\usepackage{latexsym}
\usepackage{multirow}
\usepackage{amsmath}
\usepackage{amsfonts}
\usepackage{graphicx}
\usepackage{makecell}
\graphicspath{ {/} }

\newcommand{\tabincell}[2]{\begin{tabular}{@{}#1@{}}#2\end{tabular}} 

\usepackage[T1]{fontenc}

\usepackage[utf8]{inputenc}

\usepackage{microtype}

\title{HTCInfoMax: A Global Model for Hierarchical Text Classification via Information Maximization}

\author{Zhongfen Deng\textsuperscript{1}, Hao Peng\textsuperscript{2, 3}, Dongxiao He\textsuperscript{4}, Jianxin Li\textsuperscript{2}, Philip S. Yu\textsuperscript{1}  \\ 
\textsuperscript{1}Department of Computer Science, University of Illinois at Chicago, Chicago, USA \\
  \textsuperscript{2}BDBC, Beihang University, Beijing, China \\
  \textsuperscript{3}School of Cyber Science and Technology, Beihang University, Beijing, China\\
  \textsuperscript{4}School of Computer Science and Technology, Tianjin University, Tianjin, China \\
  {\tt \{zdeng21,psyu\}@uic.edu, \{penghao,lijx\}@act.buaa.edu.cn} \\
  {\tt hedongxiao@tju.edu.cn} \\}

\def\methodname{HTCInfoMax}
\begin{document}
\maketitle
\begin{abstract}
The current state-of-the-art model HiAGM for hierarchical text classification has two limitations. 
First, it correlates each text sample with all labels in the dataset which contains irrelevant information. 
Second, it does not consider any statistical constraint on the label representations learned by the structure encoder, while constraints for representation learning are proved to be helpful in previous work. 
In this paper, we propose {\methodname} to address these issues by introducing information maximization which includes two modules: text-label mutual information maximization and label prior matching. 
The first module can model the interaction between each text sample and its ground truth labels explicitly which filters out irrelevant information. 
The second one encourages the structure encoder to learn better representations with desired characteristics for all labels which can better handle label imbalance in hierarchical text classification. 
Experimental results on two benchmark datasets demonstrate the effectiveness of the proposed {\methodname}.
\end{abstract}

\section{Introduction}
Hierarchical text classification (HTC) is a particular subtask of multi-label text classification~\cite{li2020survey}. 
Many datasets have been proposed to study HTC for decades, such as RCV1 \cite{lewis2004rcv1} and NYTimes \cite{evan2008new}, which categorize a news into several categories/labels. 
And all the labels in each dataset are usually organized as a tree or a directed acyclic graph. 
Thus, there is a label taxonomic hierarchy existing in each dataset.
The goal of HTC is to predict multiple labels in a given label hierarchy for a given text.

There are two groups of existing methods for HTC: local approaches and global approaches. 
Local approaches usually build a classifier for each label/node~\cite{banerjee2019hierarchical}, or for each parent node, or for each level of the label hierarchy\cite{wehrmann2018hierarchical,huang2019hierarchical,chang2020taming}.
Global approaches just build one classifier to simultaneously predict multiple labels of a given text. 
The earlier global approaches ignore the hierarchical structure of labels and assume there is no dependency among labels which leads to flat models such as~\cite{johnson2015effective}. 
Later on, more and more works try to make use of the label taxonomic hierarchy to improve the performance by employing different strategies such as recursively regularized Graph-CNN \cite{peng2018large}, reinforcement learning \cite{mao2019hierarchical}, attentional capsule network \cite{peng2019hierarchical}, meta-learning \cite{wu2019learning} and structure encoder \cite{zhou2020hierarchy}. 
Many attention-based models are also proposed to learn more refined text features for text classification tasks such as \cite{you2019attentionxml,deng-etal-2020-hierarchical}. 
Among these methods, HiAGM proposed by \citet{zhou2020hierarchy} is the state-of-the-art model for HTC which designs a structure encoder that integrates the label prior hierarchy knowledge to learn label representations, and then proposes a model HiAGM with two variants (one is HiAGM-LA, the other is HiAGM-TP) based on the structure encoder to capture the interactions between text features and label representations. 
However, there are some limitations of HiAGM. Firstly, it utilizes the same label hierarchy information for every text sample which cannot distinguish the relevant and irrelevant labels to a specific text sample.
Although HiAGM-LA can implicitly relate each text to its corresponding labels by soft attention weights, there are still irrelevant and noisy information. 
Secondly, for HiAGM-LA, there is no statistical constraint on the label embeddings generated by the structure encoder, while statistical constrains for representation learning are proved to be helpful by \citet{hjelm2019learning}.

To address the two limitations of HiAGM-LA, we propose {\methodname} which introduces information maximization consisting of two new modules which are text-label mutual information maximization and label prior matching on top of HiAGM-LA.
Specifically, the first new module makes a connection between each text sample and its corresponding labels explicitly by maximizing the mutual information between them, and thus can filter out irrelevant label information for a specific text sample. 
The label prior matching module can impose some constraints on the learned representation of each label to force the structure encoder to learn better representations with desirable properties for all labels and thus also improve the quality of representations for low-frequency labels, which helps handle label imbalance issue better.

In summary, our main contributions are: 
1) We propose a novel global model {\methodname} for HTC by introducing information maximization which includes two modules: text-label mutual information maximization and label prior matching. 
2) To our best knowledge, this is the first work to utilize text-label mutual information maximization for HTC which enables each text to capture its corresponding labels' information in an effective way. 
3) Also, to our best knowledge, this is the first work to introduce label prior matching for HTC which encourages the structure encoder to learn desired label representations for all labels which can better handle inherent label imbalance issue in HTC.
4) Experimental results demonstrate the effectiveness of our proposed model for HTC.
5) We release our code to enable replication, available at \url{https://github.com/RingBDStack/HTCInfoMax}.

\section{Methodology}
\subsection{Our approach}
\begin{figure*}[t]
    \centering
    \includegraphics[scale=0.59]{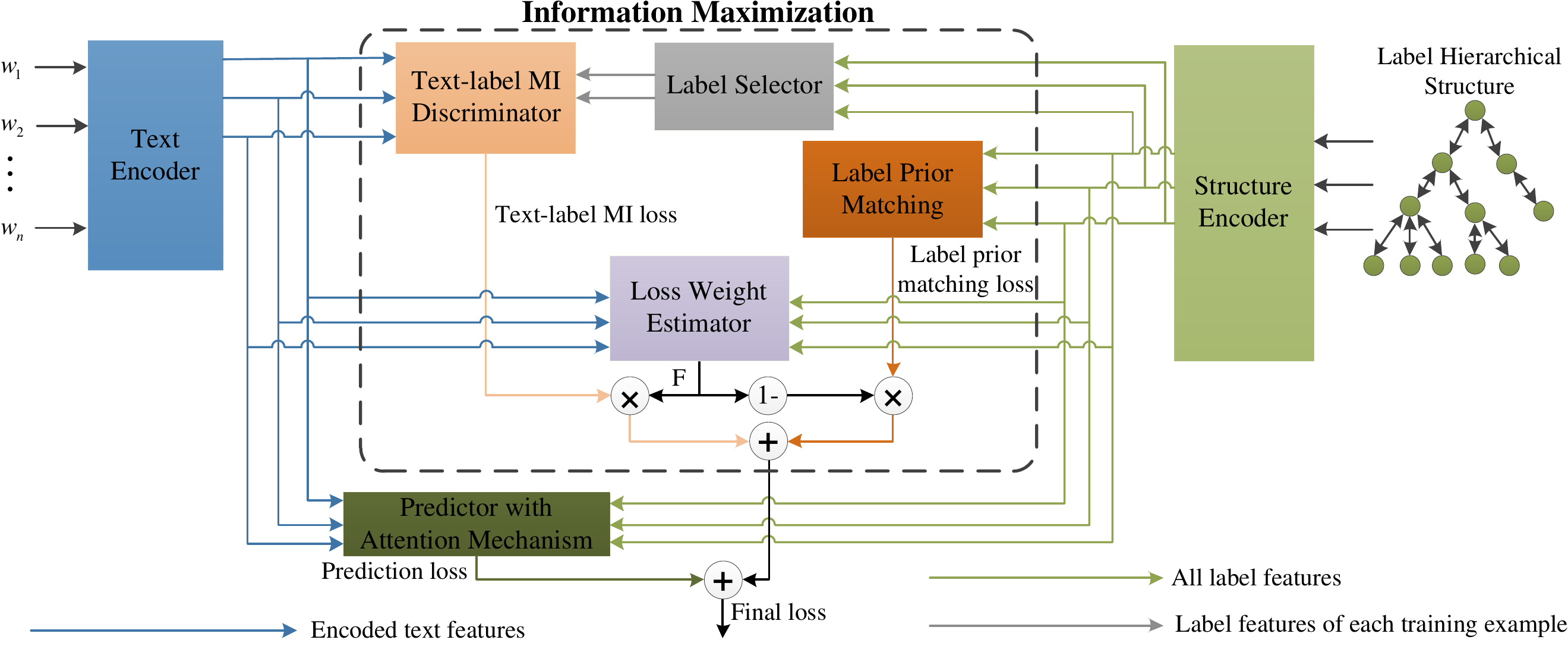}\vspace{-0.1in}
    \caption{The architecture of our model HTCInfoMax.}\vspace{-0.25in}
    \label{fig:model_architecture}
\end{figure*}

The overall architecture of our model is shown in Figure \ref{fig:model_architecture}. The major part of {\methodname} is the "Information Maximization" part shown in the dashed box which has two new modules: text-label mutual information maximization and label prior matching, which will be introduced
in the following sections. We keep the remaining part such as text encoder, structure encoder and the predictor be the same as in HiAGM-LA \cite{zhou2020hierarchy}.

\subsubsection{Text-label mutual information estimation and maximization}
Good text representation is critical for predicting its corresponding labels, thus fusing label information into text feature can help improve the prediction performance. The HiAGM-LA utilizes multi-label attention to bridge the text feature of each sample with all labels' information implicitly, which can somehow help each text obtain some label information. 
However, irrelevant label information is also injected into the text feature by using soft attention weights. 
Therefore, we design a text-label mutual information maximization module to help remove irrelevant label information for each text as well as help each text capture its corresponding labels' information. In this way, the learned representation for each text incorporates useful label information which is helpful for predicting its labels.

To implement the text-label mutual information maximization, we first select the ground truth labels for each text sample in the training process, and then apply a discriminator to estimate the mutual information between text and its labels, which is also known as negative sampling estimation. Let $\mathbb{P}_T$ and $\mathbb{P}_Y$ denote the distribution of text feature outputted by the text encoder and the distribution of label representation produced by the structure encoder respectively. And the joint distribution of text and label is denoted as $\mathbb{P}_{TY}=\mathbb{P}_{Y|T}\mathbb{P}_{T}$. 
Then the positive samples are the pairs of text $\mathbf{t}$ and its corresponding labels $\mathbf{y}$ which is denoted as $(\mathbf{t}, \mathbf{y})$, in other words, these positive samples are drawn from the joint distribution of text and label. For the negative samples, we pair $\mathbf{y}$ with another text sample $\mathbf{t'}$ in the same batch which is denoted as $(\mathbf{t'},\mathbf{y})$, the negative samples can be deemed as drawn from the product of marginal distribution of text $\mathbb{P}_T$ and label $\mathbb{P}_Y$. Both positive and negative samples are fed to the discriminator $D_{MI}$ to do classification and to estimate the mutual information $I(T;Y)$ between text and label shown in Eq. (\ref{eq:textlabelMI}). $D_{MI}(\mathbf{t,y})$ and $D_{MI}(\mathbf{t',y})$ represents the probability score assigned to the positive and negative sample by the discriminator respectively. The goal of the text-label mutual information maximization module is to maximize $I(T;Y)$, thus the loss from this module is shown in Eq. (\ref{eq:MIloss}).
\begin{equation}\label{eq:textlabelMI}
\small
\begin{split}
    I(T;Y) = &\mathbb{E}_{\mathbf{(t,y)}\sim\mathbb{P}_{TY}}[\log D_{MI}(\mathbf{t,y})] + \\ &\mathbb{E}_{\mathbf{(t',y)}\sim\mathbb{P}_{T}\mathbb{P}_{Y}}[\log(1-D_{MI}(\mathbf{t',y}))],
\end{split}
\end{equation}
\begin{equation}\label{eq:MIloss}
\small
    L_{MI} =  - I(T;Y).
\end{equation}

This module is inspired by Deep InfoMax (DIM) \cite{hjelm2019learning} which utilizes local and global mutual information maximization to help the encoder learn high-level representation for an image.
The structure of the discriminator $D_{MI}$ in this module can be found in the Appendix \ref{sec:appendix-dmi}.

\subsubsection{Label prior matching}
\label{sec:label-prior-matching}
There is an inherent label imbalance issue in HTC, thus the learned label embeddings by the model for low-frequency labels are not good because of underfitting caused by less training examples. The label prior matching imposes some statistical constrains on the learned representation of each label which can help the structure encoder learn better label representations with desirable characteristics for all labels. This also improves the quality of representations for low-frequency labels, which helps handle the label imbalance situation better in terms of improvement of Macro-F1 score.

To implement the label prior matching mechanism, we use a method similar to adversarial training in adversarial autoencoders \cite{makhzani2015adversarial} but without a generator to force the learned label representation to match a prior distribution. We denote the prior as $\mathbb{Q}$ and the distribution of label representation learned by the structure encoder as $\mathbb{P}$. Specifically, a discriminator network $D_{pr}$ is employed to distinguish the representation/sample drawn from the prior (i.e., real sample which is denoted as $\Tilde{\mathbf{y}}$) from the label embedding produced by the structure encoder (i.e., fake sample which is denoted as $\mathbf{y}$). 
For each label, we utilize $D_{pr}$ to calculate its corresponding prior matching loss $l_{pr}$ , which is shown in Eq. (\ref{eq:prloss_foreachlabel}). 
\begin{equation}\label{eq:prloss_foreachlabel}
\small
    l_{pr} = -(\mathbb{E}_{\Tilde{\mathbf{y}}\sim\mathbb{Q}}[\log D_{pr}(\Tilde{\mathbf{y}})] + \mathbb{E}_{\mathbf{y}\sim\mathbb{P}}[\log(1-D_{pr}(\mathbf{y}))]),
\end{equation}
This loss aims at pushing the distribution $\mathbb{P}$ of learned representation for a label towards its prior distribution $\mathbb{Q}$.
The final label prior matching loss is the average of losses from all the labels which is shown in Eq. (\ref{eq:laber_priormatchingloss}), $N$ is the number of labels.
\begin{equation}\label{eq:laber_priormatchingloss}
\small
    L_{pr} = \frac{1}{N} \sum_{i=1}^{N} l_{pr}^{i}.
\end{equation}

This idea is inspired by DIM which matches the representation of an image to a prior, but different from DIM, it trains the structure encoder to learn desired representations for all labels by imposing the constraints on each label's representation.

An uniform distribution on the interval [0, 1) is adopted as the label prior distribution $\mathbb{Q}$ in the label prior matching module. The reason for choosing the uniform distribution is that it works well as a prior in DIM for generating image representations. 
And the improvement of Macro-F1 score in the experimental results of hierarchical text classification further verifies the suitability of using the uniform distribution as the label prior. The detailed structure of the discriminator $D_{pr}$ can be found in the Appendix \ref{sec:appendix-dpr}.

\subsubsection{Final loss of {\methodname}}
A loss weight estimator is adopted to learn the weights for text-label mutual information loss and label prior matching loss by using learned text features $\mathbf{t}$ and all labels' representation $\mathbf{y}$, shown in Eq. (\ref{eq:lossweightesti}), and both $W_1$ and $W_2$ are trainable parameters.
\begin{equation}\label{eq:lossweightesti}
\small
    F= \operatorname{sigmoid} (W_1\mathbf{t} + W_2\mathbf{y}),
\end{equation}
And the loss from the predictor is the traditional binary cross-entropy loss $L_{c}$ \cite{zhou2020hierarchy}.
Then the final objective function of {\methodname} is the combination of all the three losses as follows:
\begin{equation}
\small
    L = L_{c} + F\times L_{MI} + (1-F)\times L_{pr}.
\end{equation}

\section{Experiment}
\subsection{Datasets and evaluation metrics}
Following HiAGM \cite{zhou2020hierarchy}, we use RCV1-V2 \cite{lewis2004rcv1} and Web of Science (WOS) \cite{kowsari2017hdltex} benchmark datasets to evaluate our model and adopt the same split of RCV1-V2 and WOS as HiAGM. 
The statistics of the two datasets are shown in Table \ref{table:dataset_statistics}.

\begin{table}[h]
\small
\centering
\begin{tabular}{p{1.35cm}<{\centering}| p{0.25cm}<{\centering} p{0.45cm}<{\centering} p{0.85cm}<{\centering} p{0.65cm}<{\centering} p{0.65cm}<{\centering} p{0.75cm}<{\centering}} 
 \hline
 Dataset & L & Depth & Avg-L & Train & Val & Test \\ [0.5ex] 
 \hline
 RCV1-V2 & 103 & 4 & 3.24 & 20,834 & 2,315 & 781,265 \\ [0.5ex]
 WOS & 141 & 2 & 2.0 & 30,070 & 7,518 & 9,397 \\ [0.5ex]
\hline
\end{tabular}
\caption{Statistics of datasets. L is the total number of labels in the dataset, Avg-L is the average number of labels for each sample. Depth means the maximum level of the label hierarchy.}
\label{table:dataset_statistics}
\end{table}

Standard evaluation metrics including Micro-F1 (Mi-F1) and Macro-F1 (Ma-F1) score are employed to evaluate our model. 
In label imbalance situation, Ma-F1 can better evaluate model's performance in the perspective of not focusing on frequent labels in a certain degree.

\subsection{Experimental setup}
In order to make a fair comparison between our model and HiAGM, we use the same parameter settings as HiAGM and follow its implementation details which can be seen in \cite{zhou2020hierarchy}. 

\subsection{Experimental results}
The experimental results of our model are shown in Table \ref{table:results_comparison}, each score is the average result of 8 runs. 
The results of HiAGM are referred from \cite{zhou2020hierarchy}.
There are two variants of HiAGM which are HiAGM-LA and HiAGM-TP. 
As stated before, our model is built on top of HiAGM-LA to address its limitations. 
From Table \ref{table:results_comparison}, one can see that our model outperforms the HiAGM-LA model with either GCN or TreeLSTM as structure encoder on two datasets, which demonstrates that the introduced information maximization in our model can address the limitations of HiAGM-LA and improve the performance.
This is because the label prior matching can drive the structure encoder to learn good and desired label representations that encode more useful and informative information of labels, and the text-label mutual information maximization module helps learn better representation of each text for prediction by fusing the above learned good representations of its ground truth labels while ignoring irrelevant labels' information. 
It is also worth nothing that the improvement of Ma-F1 on the RCV1-V2 dataset is bigger compared with that on WOS, which indicates that our model can work better on dataset with a more complicated label hierarchy as RCV1-V2 has a deeper label hierarchical structure than WOS. 

\begin{table}[h]
\small
\centering
\begin{tabular}{p{1.09cm}<{\centering} p{0.98cm}<{\centering}| p{0.85cm}<{\centering} p{0.87cm}<{\centering}| p{0.85cm}<{\centering} p{0.87cm}<{\centering}} 
 \hline
		\multicolumn{2}{c|}{\multirow{2}{*}{Models}} &\multicolumn{2}{c|}{RCV1-V2} & \multicolumn{2}{c}{WOS} \\ \cline{3-6} 
		& & Mi-F1 & Ma-F1 & Mi-F1 & Ma-F1 \\ [0.5ex] 
 \hline
 \multirow{2}{1.38cm}{{\tabincell{l}{HiAGM-LA}}}
 & \tabincell{l}{\tiny GCN} & 82.21 & 61.65 & 84.61 & 79.37 \\ [0.5ex]
 & \tabincell{l}{\tiny TreeLSTM} & 82.54 & 61.90 & 84.82 & 79.51 \\ [0.5ex]
 \hline
 \multirow{2}{1.38cm}{{\tabincell{l}{HiAGM-TP}}}
 & {\tabincell{l}{\tiny GCN}} & \underline{\emph{\textbf{83.96}}} & \underline{\textit{\textbf{63.35}}} & \underline{\textit{\textbf{85.82}}} & \underline{\textit{\textbf{80.28}}} \\ [0.5ex]
 & {\tabincell{l}{\tiny TreeLSTM}} & 83.20 & 62.32 & 85.18 & 79.95 \\ [0.5ex]
 \hline
 \multicolumn{2}{l|}{{\methodname} (Ours)} & \textbf{83.51} & \textbf{62.71} & \textbf{85.58} & \textbf{80.05} \\ [0.5ex]
\hline
\end{tabular}
\caption{Results of {\methodname} and HiAGM on RCV1-V2 and WOS datasets.}
\label{table:results_comparison}
\end{table}

Although our model does not outperform all the results of HiAGM-TP, it reaches the similar performance.
This indicates that information maximization is an alternative effective way to fuse the text feature and label information together to boost the performance. 
In addition, apart from generating text representations, our model can also generate refined label representations via information maximization which can be utilized for inference, while HiAGM-TP cannot produce such label embeddings for usage in the inference phase because it directly feeds the text feature into the structure encoder to obtain final text representation for prediction.
In other words, HiAGM-TP encodes text and label information into only one feature space. However, obtaining separate text features and label features such as the ones generated by our model can help encode more semantic information of labels, which may be helpful for HTC especially when there is a large label hierarchy in the dataset.

We do not report the results of other baselines such as HFT(M) \cite{shimura-etal-2018-hft}, 
SGM \cite{yang-etal-2018-sgm}, 
HiLAP-RL \cite{mao2019hierarchical}, etc. as they can be found in \cite{zhou2020hierarchy}, and our model performs better than these baselines.

\subsection{Ablation study}
To demonstrate the effectiveness of the two modules of information maximization, we conduct an ablation study and the results are shown in Table \ref{table:results_ablation_study}. Every score in Table \ref{table:results_ablation_study} is the average result of 8 runs. 
From Table \ref{table:results_ablation_study}, one can see that {\methodname} outperforms the variant without text-label mutual information maximization module (i.e., {\methodname} w/o MI) by 0.09, 0.92 points on RCV1-V2 and 0.12, 0.11 points on WOS in terms of Mi-F1 and Ma-F1 respectively, which indicates that the text-label mutual information maximization module can make each text capture its corresponding labels' information and thus improves the Mi-F1 and Ma-F1 score at the same time. 
When compared with the other variant (i.e., {\methodname} w/o LabelPrior), the improvements of the two metrics can also be observed but Ma-F1 has larger improvements by 2.14 and 1.04 points on RCV1-V2 and WOS respectively compared with Mi-F1. This demonstrates that label prior matching helps regularize the label feature space and forces the structure encoder to learn better representations with desired properties for all labels. Thus the representations of imbalanced labels are also well learned, which helps mitigate the issue of underfitting of low-frequency labels, and thus improves the Ma-F1 score more and better to handle the label imbalance issue.

\begin{table}[t]
\small
\centering
\begin{tabular}{m{2.0cm}<{\centering}| m{0.95cm}<{\centering} m{0.95cm}<{\centering}| m{0.95cm}<{\centering} m{0.95cm}<{\centering}} 
 \hline
		\multirow{2}{*}{Models}    &\multicolumn{2}{c|}{RCV1-V2} & \multicolumn{2}{c}{WOS} \\ \cline{2-5} 
		& Mi-F1 & Ma-F1 & Mi-F1 & Ma-F1 \\ [0.5ex] 
 \hline
 {\methodname} w/o MI & 83.42 & 61.79 & 85.46 & 79.94 \\ [0.5ex]
 \hline
 {\methodname} w/o LabelPrior & 82.75 & 60.57 & 84.74 & 79.01 \\ [0.5ex]
 \hline
 {\methodname} & \textbf{83.51} $\uparrow$ & \textbf{62.71} $\uparrow$ & \textbf{85.58} $\uparrow$ & \textbf{80.05} $\uparrow$ \\ [0.5ex]
\hline
\end{tabular}
\caption{Ablation study results on RCV1-V2 and WOS datasets. w/o means without. Arrow $\uparrow$ indicates statistical significance ($p < 0.01$).}
\label{table:results_ablation_study}
\end{table}

\section{Conclusion}
We propose {\methodname} to address the limitations of HiAGM by introducing information maximization which includes two modules: text-label mutual information maximization and label prior matching. The label prior matching can drive the model to learn better representations for all labels, while the other module further fuses such learned label representations into text to learn better text representations containing effective label information for prediction. The experimental results demonstrate the effectiveness of {\methodname}.

\section*{Acknowledgment}
The corresponding author is Hao Peng.
The authors of this paper were supported by NSFC through grants U20B2053, 62002007, 62073012 and 61876128,
S\&T Program of Hebei through grant 20310101D,
and in part by NSF under grants III-1763325, III-1909323, and SaTC-1930941. 
We thank the reviewers for their constructive feedback.
\bibliographystyle{acl_natbib}
\bibliography{main}

\appendix

\section{Architecture Details of Information Maximization}
\label{sec:appendix}

\subsection{The structure of discriminator in text-label mutual information maximization module}
\label{sec:appendix-dmi}
\begin{figure}[h]
    \centering
    \includegraphics[scale=0.55]{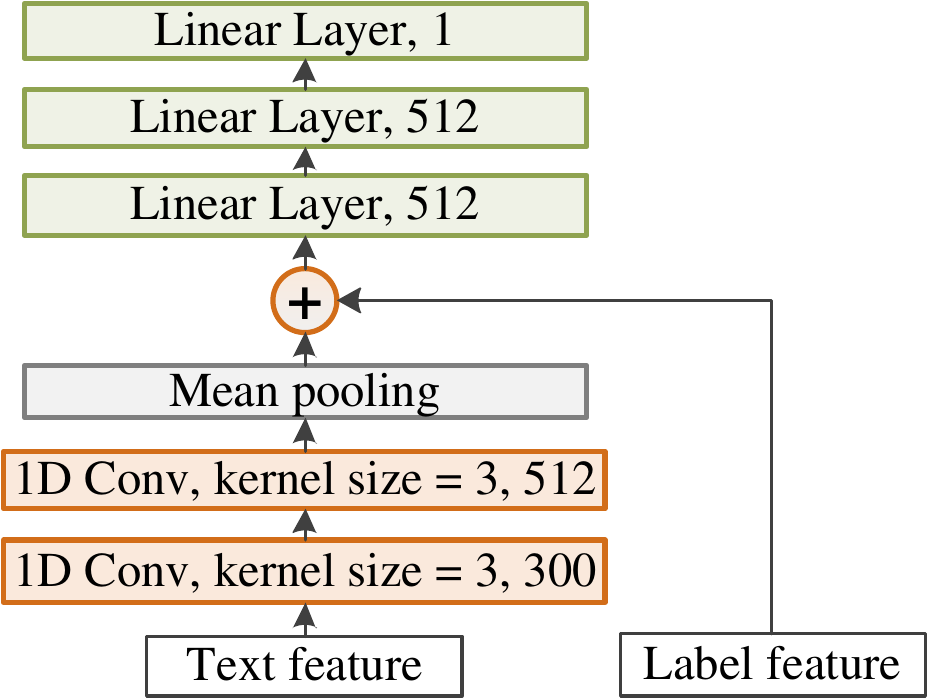}\vspace{-0.1in}
    \caption{The structure of discriminator $D_{MI}$.}\vspace{-0.25in}
    \label{fig:dmi_structure}
\end{figure}

The discriminator $D_{MI}$ consists of two 1D-convolutional layers with kernels of size 3 and three linear layers. The architecture of $D_{MI}$ is shown in Figure \ref{fig:dmi_structure} and the details of all the layers are shown in Table \ref{table:dmi_disc} ("-" indicates that there is no activation for the corresponding layer). As shown in Figure \ref{fig:dmi_structure}, the discriminator $D_{MI}$ takes pairs of text representation and label representation as input. The text representations are fed to the convolutional layers first, then the label representations are concatenated with the output from the convolutional layers and fed to the following linear layers. The final linear layer produces a score for each pair of text sample and corresponding labels.

\begin{table}[h]
\small
\centering
\begin{tabular}{c| c c c} 
 \hline
 Layers & Size (Input) & Size (Output) & Activation \\ [0.5ex] 
 \hline
 1D-conv layer & 300 & 300 & ReLU \\ [0.5ex]
 1D-conv layer & 300 & 512 & - \\ [0.5ex]
 Linear layer & 812 & 512 & ReLU \\ [0.5ex]
 Linear layer & 512 & 512 & ReLU \\ [0.5ex]
 Linear layer & 512 & 1 & - \\ [0.5ex]
\hline
\end{tabular}
\caption{Layer details of the discriminator $D_{MI}$.}
\label{table:dmi_disc}
\end{table}

\subsection{The structure of discriminator in label prior matching}
\label{sec:appendix-dpr}
The discriminator $D_{pr}$ in the label prior matching module is composed of three linear layers. The details of these layers are shown in Table \ref{table:dpr_disc}. This discriminator takes label representation as input and is applied for each label to compute its prior matching loss as stated in Section \ref{sec:label-prior-matching}.

\begin{table}[h]
\small
\centering
\begin{tabular}{c| c c c} 
 \hline
 Layers & Size (Input) & Size (Output) & Activation \\ [0.5ex] 
 \hline
 Linear layer & 300 & 1000 & ReLU \\ [0.5ex]
 Linear layer & 1000 & 200 & ReLU \\ [0.5ex]
 Linear layer & 200 & 1 & Sigmoid \\ [0.5ex]
\hline
\end{tabular}
\caption{Layer details of the discriminator $D_{pr}$.}
\label{table:dpr_disc}
\end{table}

\end{document}